\documentclass[10pt,twocolumn,letterpaper]{article}

\usepackage{cvpr}              

\usepackage[dvipsnames]{xcolor}

\definecolor{cvprblue}{rgb}{0.21,0.49,0.74}
\usepackage[pagebackref,breaklinks,colorlinks,citecolor=cvprblue]{hyperref}

\def\paperID{4318} 
\def\confName{CVPR}
\def\confYear{2024}

\usepackage{graphicx}
\usepackage{amsmath}
\usepackage{amssymb}
\usepackage{booktabs}
\usepackage{multirow}
\usepackage[accsupp]{axessibility}

\title{ID-Blau: Image Deblurring by Implicit Diffusion-based reBLurring AUgmentation}

\author{{{Jia-Hao Wu{$^{*, \textcolor{red}{1}}$}}\;\;
{Fu-Jen Tsai{$^{*, \textcolor{red}{2}}$}}\;\;
{Yan-Tsung Peng$^{\textcolor{red}{3}}$}\;\;
{Chung-Chi Tsai$^{\textcolor{red}{4}}$}\;\;
{Chia-Wen Lin$^{\textcolor{red}{2}}$}\;\;
{Yen-Yu Lin$^{\textcolor{red}{1}}$}}
\\
{{National Yang Ming Chiao Tung University$^{\textcolor{red}{1}}$}\;\;
{National Tsing Hua University$^{\textcolor{red}{2}}$}}
\\
{{National Chengchi University$^{\textcolor{red}{3}}$}\;\;
{Qualcomm Technologies, Inc.$^{\textcolor{red}{4}}$}}
\\
{\tt\small {jiahao.11@nycu.edu.tw \;\; fjtsi@gapp.nthu.edu.tw \;\; ytpeng@cs.nccu.edu.tw}}
\\
{\tt\small {charles0184@gmail.com \;\; cwlin@ee.nthu.edu.tw \;\; lin@cs.nycu.edu.tw}}
}
\begin{document}
\maketitle
\renewcommand{\thefootnote}{}
\footnotetext{\tt\small$*$ equal contribution}
\begin{abstract}
Image deblurring aims to remove undesired blurs from an image captured in a dynamic scene. Much research has been dedicated to improving deblurring performance through model architectural designs. However, there is little work on data augmentation for image deblurring.
Since continuous motion causes blurred artifacts during image exposure, we aspire to develop a groundbreaking blur augmentation method to generate diverse blurred images by simulating motion trajectories in a continuous space.
This paper proposes Implicit Diffusion-based reBLurring AUgmentation (ID-Blau), utilizing a sharp image paired with a controllable blur condition map to produce a corresponding blurred image.
We parameterize the blur patterns of a blurred image with their orientations and magnitudes as a pixel-wise blur condition map to simulate motion trajectories and implicitly represent them in a continuous space.
By sampling diverse blur conditions, ID-Blau can generate various blurred images unseen in the training set.
Experimental results demonstrate that ID-Blau can produce realistic blurred images for training and thus significantly improve performance for state-of-the-art deblurring models.
The source code is available at \href{https://github.com/plusgood-steven/ID-Blau}{https://github.com/plusgood-steven/ID-Blau}.

\end{abstract}
\section{Introduction}

Camera shake or object movements cause unpleasant motion blurs when we capture images.
Such blurs are usually non-uniform, leading to locally and globally undesirable artifacts. Image deblurring aims to restore the sharpness of a blurred image, a highly ill-posed problem that has remained challenging over the past decades.

Image deblurring has reached remarkable progress with the rise of deep learning. Numerous methods based on convolutional neural networks (CNNs) have rendered success in deblurring. These methods mostly adopted recurrent-based architectures, such as multi-scales~\cite{Nah_2017_CVPR, tao2018srndeblur, gao2019dynamic}, multi-patches~\cite{Zhang_2019_CVPR, Zamir2021MPRNet}, and multi-temporal~\cite{MT_2020_ECCV}. 
In addition to CNN-based methods, several studies have shown significant improvement using Transformers~\cite{IPT, Zamir2021Restormer, Tsai2022Stripformer, Wang_2022_CVPR, Kong_2023_CVPR}. Transformers~\cite{vaswani2017attention} utilize self-attention mechanisms to extract longer-range features than CNNs but require more memory. To overcome the high memory usage in the vanilla self-attention mechanism, some efficient attention mechanisms have been proposed, such as the channel-wise~\cite{Zamir2021Restormer} and strip-wise attention~\cite{Tsai2022Stripformer} mechanisms.

\begin{figure}
\begin{center}
\includegraphics[width=1\columnwidth]{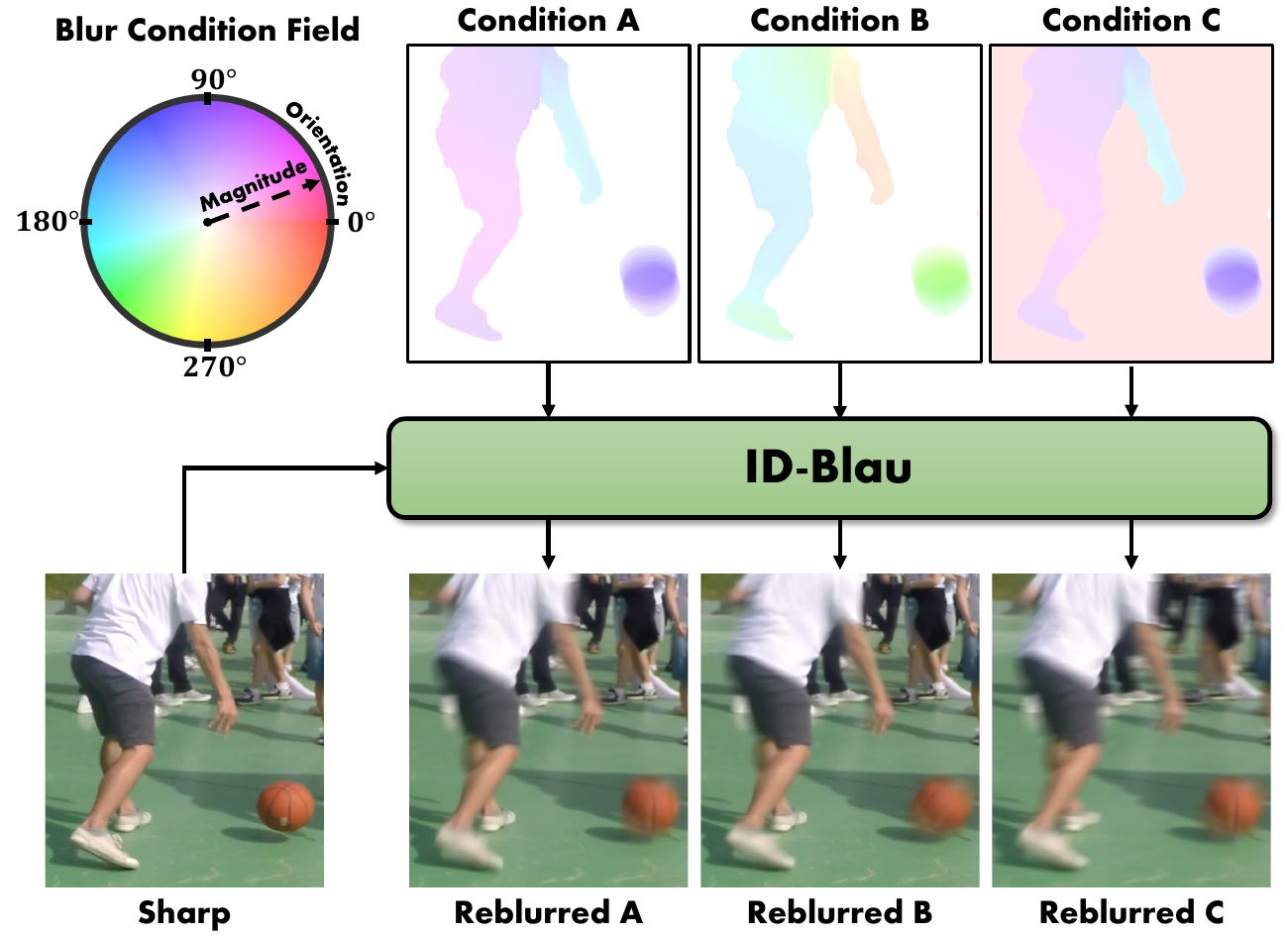}
\end{center}
\vspace{-0.2in}
\caption{{Examples of continuous reblurring by ID-Blau, where blur condition maps represent pixel-wise blur information, consisting of blur orientations and magnitudes, in a continuous space. ID-Blau can take a sharp image and a blur condition map to synthesize a blurred image, even unseen in the training set. Condition A is a blur condition map computed from the GoPro training set, which can be used to reblur a sharp image to generate a blurred image as provided in the training set. We can create Condition B and C based on A to synthesize new reblurred images, where Condition B is Condition A with rotated orientations, and Condition C is Condition A added with a camera motion blur.}
}
\label{fig:introduction}
\vspace{-0.20in}
\end{figure}

Although previous methods have improved deblurring performance through architectural designs, an effective data augmentation strategy for image deblurring has been less studied.  
A common strategy~\cite{zhang2020deblurring} is to synthesize blurred images through Generative Adversarial Networks (GANs)~\cite{NIPS2014_5ca3e9b1}. However, GANs usually generate uncontrollable results~\cite{yu2020inclusive, Shoshan_2021_ICCV, KowalskiECCV2020}, constraining the potential of blur augmentation strategies.  
As a result, we resort to developing a stable and controllable blur augmentation method based on the inductive bias of motion blurs. 
Since continuous motions cause blur patterns during exposure, we attempt to implicitly characterize blurs, such as blur orientations and magnitudes, in a continuous space where we can sample blur trajectories to generate various blurred images.   
%
%
%
This enables us to generate various blurred images with specified blur conditions, devising a stable and controllable augmentation strategy for image deblurring.     

In this paper, we propose \textbf{I}mplicit \textbf{D}iffusion-based re\textbf{BL}urring
\textbf{AU}gmentation (\textbf{ID-Blau}) that can produce diverse blur and sharp training pairs for training to improve image deblurring performance. ID-Blau generates a realistic blurred image from a sharp image and a  blur condition map to control the blur patterns through pixel-wise blur condition maps. Several examples are shown in Figure~\ref{fig:introduction}.
To create blur condition maps, we implicitly represent blur information, including blur orientations and magnitudes, in a continuous space, where we can sample various blur conditions to simulate real blur trajectories during exposure.    
Compared to the GoPro training set~\cite{Nah_2017_CVPR}, which accumulates consecutive sharp frames to generate explicit blurred images, our strategy can generate various blurred images with sampled blur conditions in the continuous blur conditoin field.
In addition, motivated by the remarkable generation ability of diffusion models~\cite{NEURIPS2020_4c5bcfec}, ID-Blau adopts diffusion models to take implicit blur condition maps to generate realistic and controllable blurred images.   
Our experimental results demonstrate that ID-Blau, the proposed implicitly conditional diffusion model, can generate diverse blur and sharp training pairs for training, even for those unseen in the training set. As a result, it significantly improves the performance of existing deblurring models and performs favorably against state-of-the-art augmentation methods in this regard. Our contributions are summarized as follows:
\begin{itemize}
\vspace{0.1in}
  \item We propose ID-Blau, a stable and controllable blur augmentation strategy for enhancing dynamic scene image deblurring.
  \vspace{0.15in}
  \item We model a continuous blur condition field to implicitly represent blur orientations and magnitudes, where we can sample various pixel-wise blur condition maps to generate diverse reblurred images not provided in the training set.  
  \vspace{0.15in}
  \item The proposed ID-Blau integrates pixel-wise blur condition maps into a diffusion model to generate high-quality reblurred images.    
  \vspace{0.15in}
  \item Experimental results show that ID-Blau significantly improves existing deblurring models and performs favorably against state-of-the-art deblurring methods.

\end{itemize}

\section{Related Work}
\subsection{Image Deblurring} 
The development of CNNs has advanced image deblurring remarkably. Several studies improved deblurring performance through recurrent networks, such as scale-recurrent~\cite{Nah_2017_CVPR, tao2018srndeblur} and patch-recurrent~\cite{Zhang_2019_CVPR,Zamir2021MPRNet} networks. 
Specifically, Nah \etal~\cite{Nah_2017_CVPR} designed a scale-recurrent method to deal with image blur in a coarse-to-fine manner. Zhang \etal~\cite{Zhang_2019_CVPR} developed a patch-recurrent method to leverage local information in a hierarchical deblurring network. 
Other than recurrent designs, Cho \etal~\cite{MIMO} proposed an efficient deblurring model that rethinks the coarse-to-fine strategy in a single-forward model to reduce latency.
Recently, motivated by the success of Vision Transformer (ViT)~\cite{dosovitskiy2020vit}, several works~\cite{Zamir2021Restormer,Tsai2022Stripformer,Kong_2023_CVPR} resorted to Transformer-based architectures to deblur an image. 
For example, Zamir \etal~\cite{Zamir2021Restormer} utilized channel-wise attention to alleviate the tremendous memory load in the vanilla transformer for self-attention. Tsai \etal~\cite{Tsai2022Stripformer} proposed strip attention that exploits strip features to better extract blur features with various orientations and magnitudes. Kong \etal~\cite{Kong_2023_CVPR} proposed a frequency-based attention mechanism that utilizes element-wise multiplication in the frequency domain to replace the dot product in the spatial domain.

\subsection{Deblurring through Reblurring} 
Besides focusing on architectural designs, some works~\cite{zhang2020deblurring,nah2021clean,chi2021test,Liu_2022_BMVC} enhanced deblurring performance through reblurring models. Chi \etal~\cite{chi2021test} and Liu \etal~\cite{Liu_2022_BMVC} utilized a reblurring network in meta-learning to achieve test-time adaptation.
Nah \etal~\cite{nah2021clean} proposed a reblurring network to amplify the blurs in a deblurred image while keeping sharp parts unchanged. It can be used to reinforce a deblurring network to generate a sharp result more accurately. 
Another potential strategy to improve deblurring models is to augment training pairs through reblurring models. Zhang \etal~\cite{zhang2020deblurring} exploited real-world blurry images to synthesize blurred images through GANs. 
However, GANs usually generate uncontrollable results, and GANs' unstable optimization process often increases the difficulty of training~\cite{8237566, NIPS2017_892c3b1c, pmlr-v70-arjovsky17a}. Besides, GANs' poor mode convergence~\cite{xiao2022DDGAN,thanh-tung2018improving} makes them less effective in synthesizing diverse blurred images.
In contrast, we propose to represent blur conditions in a continuous space implicitly and adopt a diffusion method to generate diverse, realistic, and high-quality blurry images by considering arbitrarily enumerated blur conditions.

\subsection{Diffusion Models} 
Diffusion models~\cite{NEURIPS2020_4c5bcfec,song2020denoising,Rombach_2022_CVPR} have demonstrated their outstanding ability for image synthesis. Unlike GANs, diffusion models have a stable training strategy by synthesizing images through forward diffusion and backward denoising processes. The former gradually adds Gaussian noise to a clear image in multiple steps and generates noisy images in a sequence. The latter iteratively denoises the degraded images to restore the original image.
Other than image synthesis, several studies have successfully applied diffusion models to low-level vision tasks, such as super-resolution~\cite{saharia2021image, Gao_2023_CVPR, yue2023resshift}, inpainting~\cite{Lugmayr_2022_CVPR, Xie_2023_CVPR, Anciukevicius_2023_CVPR}, and deblurring~\cite{Whang_2022_CVPR, Ren_2023_ICCV, chen2023hierarchical}.
Specifically, Saharia \etal~\cite{saharia2021image} conditioned the diffusion models on low-resolution images to reconstruct their high-resolution versions with visually pleasant quality. Whang \etal~\cite{Whang_2022_CVPR} conditioned the diffusion models on deterministic deblurring results to improve perceptual quality.  
Instead of constructing a restoration network based on diffusion models, we propose an implicitly conditional diffusion reblurring model to generate high-quality training data for improving deblurring performance for existing models.
We can generate various blur conditions in a continuous space implicitly to control the proposed reblurring model to synthesize blurred images consistent with these conditions. By sampling the blur condition space, a set of diverse and high-quality blurred images can be generated to enrich the training set and enhance the performance of existing deblurring models.

\section{Proposed Method}

This section presents the proposed ID-Blau, which turns a sharp image into a blurred version based on a pixel-wise blur condition map. 
We characterize blurs with their orientations and magnitudes in a continuous space, where we can sample blur conditions for reblurring a sharp image.
With the modeling, we can simulate blur trajectories yielded by continuous motion during exposure. 
It allows us to implicitly manipulate blur conditions to generate diverse blurred images unseen during training for data augmentation.
In the data augmentation scenario, we can use ID-Blau to produce additional training data before optimizing deblurring models, so we do not need to run ID-Blau when optimizing deblurring models.
The following details the computation of blur conditions and the training and sampling processes of ID-Blau for blurred image generation.

%

\subsection{Blur Conditions}
During exposure, continuous motion from a camera or scene objects may cause blur artifacts to captured images. 
Thus, global and/or local blur trajectories can be found in such blurred images. 
To synthesize blurred images from their sharp counterparts, Nah \etal~\cite{Nah_2017_CVPR} proposed the GoPro dataset, approximating the continuous exposure through a blur accumulation function as
\vspace{-0.05in}
\begin{equation}
    B = g(\frac{1}{T}\int_{t=1}^{T}V(t)dt) \simeq g(\frac{1}{N}\sum_{n=1}^{N}V\left[n\right]),
\label{eqa:accumulaiton}
\end{equation}
where $B$, $V(t)$, $T$, $N$, and $g$ denote the generated blurred frame, sharp frame at time $t$ collected by a high-speed camera, exposure time, number of sampled sharp frames, and camera response function, respectively. 
The accumulation function aggregates a sharp image sequence $\textbf{V}=\{V\left[1 \right], ..., V\left[N \right]\}$ to generate one blurred image $B$. 
The center sharp frame $V[\frac{N+1}{2}]$ is assigned to the ground-truth image $S$ for $B$, and $N$ is typically an odd number.

Continuous motions of different objects may cause various blur patterns since they can move independently. 
We propose to characterize the blurs in an image into a blur condition map specifying a blur orientation and magnitude to each pixel. 
The blur condition map depicts a specific blur scenario for the blurred image by vectorizing blur for each pixel in a continuous field.
By changing the blur condition map, we can simulate various blur scenarios and augment diverse blurred images to complement existing image deblurring datasets.
To obtain the corresponding blur condition from a blurred image, we choose to average bidirectional optical flows from the sharp sequence $\textbf{V}$ and then aggregate them by summation as 
\begin{equation}
    \mathcal{F} = \sum_{n=1}^{N-1}\frac{f_\theta(V\left[n\right], V\left[n+1\right]) - f_\theta(V\left[n+1\right], V\left[n\right])}{2}, 
\end{equation}
where $f_\theta$ denotes the pre-trained optical flow estimation network~\cite{RAFT}, $\mathcal{F}=[u; v] \in\mathbb{R}^{H \times W \times 2}$ denotes the overall motion trajectories during exposure, where $H$ and $W$ are the image height and width, respectively.
The tensor $\mathcal{F}$ records the horizontal and vertical motion trajectories in $u$ and $v\in\mathbb{R}^{H \times W}$.
%

\begin{figure}[t!]
\begin{center}
\includegraphics[width=1\columnwidth]{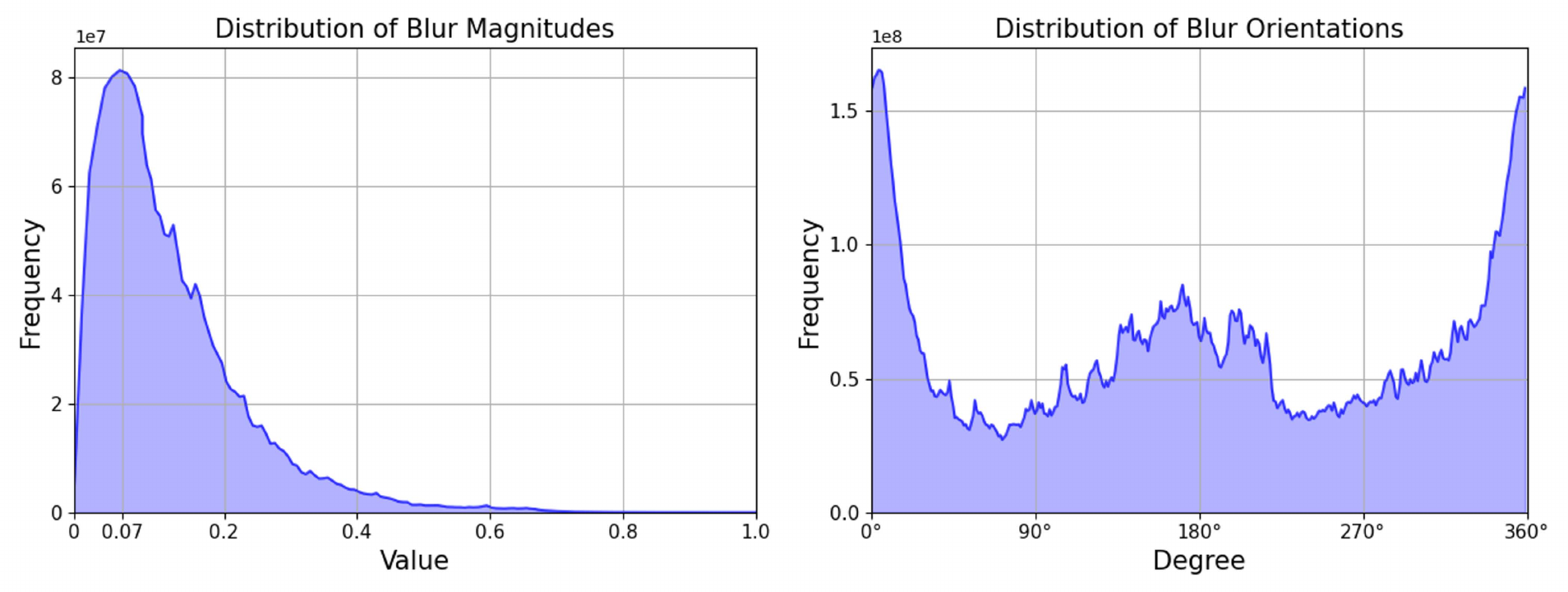}
\end{center}
\vspace{-0.3in}
\caption{Illustration of distributions of blur magnitudes (left) and orientations (right)
of the GoPro training set.
}
\label{fig:condition_space}
\vspace{-0.15in}
\end{figure}

\begin{figure*}[t!]
\begin{center}
\includegraphics[width=1\textwidth]{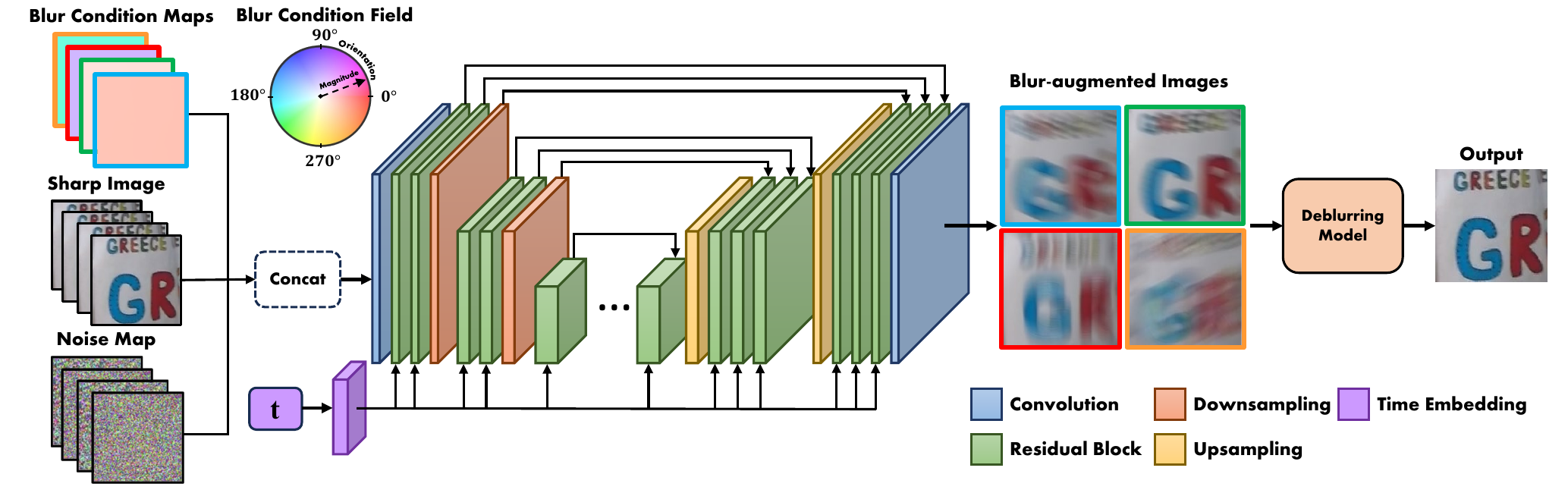}
\end{center}
\vspace{-0.2in}
\caption{Reblurring process with ID-Blau. As the pie chart shows the blur condition field with orientations and magnitudes delineated in different colors, we visualize the process of generating blurred images with a set of blur condition maps. A sharp image paired with a blur condition map and a noise map is concatenated and fed into ID-Blau to produce a blurred image, where an MLP is used to encode the iteration index $t$ as Time Embedding. Using ID-Blau can augment an image deblurring training set offline for optimizing a deblurring model and improving its performance.
}
\label{fig:framework}
\vspace{-0.2in}
\end{figure*}

Next, we normalize $\mathcal{F}$ to obtain unit motion vectors and append one dimension of the magnitudes to make it a 3D tensor $C=[x;y;z] \in\mathbb{R}^{H \times W \times 3}$, where $x_{i,j}=\frac{u_{i,j}}{\sqrt{u_{i,j}^2 + v_{i,j}^2}}$ and $y_{i,j}=\frac{v_{i,j}}{\sqrt{u_{i,j}^2 + v_{i,j}^2}}$ respectively represent the horizontal and vertical blur orientations for a pixel $(i,j)$, and $z_{i,j}=\frac{\sqrt{u_{i,j}^2 + v_{i,j}^2}}{M}$ denotes its corresponding blur magnitudes. 
Here, $M$ is set to the largest blur magnitude in a set of considered blur conditions to make $z \in [0,1]$.
%

Let $C$ be the blur condition map for a blurred image $B$ in a normalized continuous space, corresponding to its ground-truth image $S$.
With the above modeling, we can construct a set of $K$ blurred and sharp image pairs with their computed blur condition maps $\{B_n, S_n, C_n\}^{K}_{n=1}$ to train a reblurring model. 
It uses a sharp image $S_n$ and a blur condition map $C_n$ to produce the corresponding blurred image $B_n$, correlating the sharp image $S_n$ and its blurred version $B_n$ with our computed blur condition map $C_n$. 

For example, we show the distribution of the blur magnitude and orientation of GoPro in Figure~\ref{fig:condition_space}, indicating that the magnitude histogram peaks towards the left side, meaning most images are not heavily blurred. 
In contrast, the orientation has slight peaks around 0, 180, and 360 degrees, implying more blurs tend to be horizontal, but cases in other orientations can also be found.
Training our reblurring model on such diverse data can correlate a blurred image with its computed blur information implicitly in a continuous blur condition field and empower the model to generate blurred images unseen in the original data by sampling different blur magnitudes and orientations.
In the following, we describe optimizing the proposed diffusion-based reblurring model in ID-Blau.



\subsection{ID-Blau}
ID-Blau is developed based on a conditional diffusion model that takes a blur condition map $C$ and a sharp image $S$ as input to generate a blurred image $B$, as shown in Figure~\ref{fig:framework}. 
Therefore, we can create various blur conditions to augment data with realistic reblurred images unseen in the training set.
To expedite the inference, we adopt the denoising diffusion implicit model (DDIM)~\cite{song2020denoising}, utilizing the same training procedure as DDPM~\cite{NEURIPS2020_4c5bcfec} but accelerating the inference via a non-Markovian inference process.

\begin{figure*}[t!]
\begin{center}
\includegraphics[width=1\textwidth]{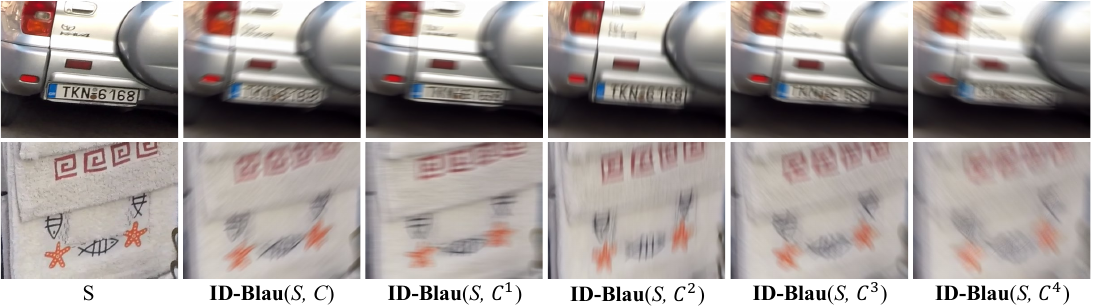}
\end{center}
\vspace{-0.25in}
\caption{Illustration of reblurred images with ID-Blau. It takes a sharp image $S$ and a blur condition map $C=[x; y; z]$ as inputs to generate the corresponding blurred image. We show several generated blurred images by altering $C$, such as unit horizontal or vertical blur orientations, $C^1=[1; 0; z]$ and $C^2=[0; 1; z]$, horizontally inverse $C$ as $C^3=[-x; y; z]$, and $C^4=[-x; y; 2z]$ with twice the magnitude for $C^3$.   
}
\label{fig:method_viz1}
\vspace{-0.1in}
\end{figure*}


\vspace{-0.2in}
\paragraph{Training Stage.} 
Following DDPM~\cite{NEURIPS2020_4c5bcfec}, we perform a forward diffusion process and a backward denoising process to optimize ID-Blau.  
In the forward diffusion process, we gradually add Gaussian noise to a blurred image $B$ over $T$ iterations for generating a sequence of noisy images $\{B_1, ..., B_T\}$, where $T$ is set to $2,000$. 
The sampling probability $q(B_{1:T}|B_{0}) $ is defined as
\begin{equation}
\begin{split}
q(B_{1:T}|B_{0}) &= \prod_{t=1}^{T}q(B_{t}|B_{t-1});\\
q(B_t|B_{t-1}) &= \mathcal{N}(B_t;\sqrt{\alpha_t}B_{t-1},(1-\alpha_t)\mathrm{I}),    
\end{split}
\label{eqa:forward}
\end{equation}
where $B_{0}=B$ and $\{\alpha_t\in(0, 1)\}$ is a set of hyperparameters that controls the noise variance over a sequence of steps. 
Instead of iteratively adding Gaussian noise to $B$, the forward diffusion process can be reparameterized  as 
\begin{equation}
B_t = \sqrt{\overline{\alpha}_t}B_0 + (1-\sqrt{\overline{\alpha}_t})\epsilon,
\end{equation}
where $\overline{\alpha}_t=\prod_{i=1}^{t}\alpha_i$ and $\epsilon\sim\mathcal{N}(0,\mathrm{I})$. 

In the backward denoising process, the blurred image $B$ is iteratively restored from the noisy image $B_T$ through a denoising model $\epsilon_\theta$. 
When recreating a blurred image $B$ from a noisy image, we condition the denoising model on its sharp counterpart $S$ and the blur condition map $C$ computed based on $B$.
Thus, the denoising model learns to correlate a blurred image $B$ and its sharp image $S$ conditioned on the blur condition map $C$. The reverse conditional probability is written as
\begin{equation}
\begin{split}
&p_\theta(B_{t-1}|B_t, S, C) = \mathcal{N}(B_{t-1};\mu_\theta(B_t, S, C, t),\sigma^2_t\mathrm{I});\\    
&\mu_\theta(B_t, S, C, t)=\frac{1}{\sqrt{\alpha_t}}(B_t - \frac{1 - \alpha_t}{\sqrt{1-\overline{\alpha}_t}}\epsilon_\theta(B_t, S, C, t)),
\end{split}
\end{equation}
where the Gaussian density function is parameterized with the mean of $\mu_\theta(B_t, S, C, t)$ and variance of $\sigma^2_t = (1 - \alpha_t)$.
To train the denoising model, we use the following objective function:
\begin{equation}
\mathcal{L} = \parallel\epsilon-\epsilon_\theta(\sqrt{\overline{\alpha}_t}B_0 + \sqrt{(1-\overline{\alpha}_t)}\epsilon, S, C,t)\parallel_1,
\end{equation}
where $\sqrt{\overline{\alpha}_t}B_0 + \sqrt{(1-\overline{\alpha}_t)}\epsilon$ is the noisy image $B_t$ at the $t$-th step. Figure~\ref{fig:framework} shows the architecture of the denoising model $\epsilon_\theta$.

\vspace{-0.2in}
\paragraph{Inference Stage.} 
After optimizing the denoising model, ID-Blau can generate a blurred image $B$ from a pure noisy image conforming to a Gaussian distribution $B_T\sim\mathcal{N}(0,\mathrm{I})$, a sharp image $S$, and a blur condition map $C$ through the sampling procedure in DDIM. 
It can take any given blur condition map implicitly created to produce diverse blurred images unseen in the training set. The denoising process is given by:
%
\begin{equation}
\begin{split}
B_{t-1} = & \sqrt{\overline{\alpha}_{t-1}}(\frac{B_t - \sqrt{1-\overline{\alpha}_t}\epsilon_\theta(B_t, S, C, t)}{\sqrt{\overline{\alpha}_t}}) \\ 
& + \sqrt{1-\overline{\alpha}_{t-1}-\sigma^2_t}\cdot\epsilon_\theta(B_t, S, C, t)+\sigma_t z,
\end{split}
\end{equation}
where $\sigma_t = 0$, making the process deterministic. Through DDIM, we set the number of iterations during sampling to $20$ without excessively sacrificing the image generation quality. 

Figure~\ref{fig:method_viz1} shows examples of generating different blurred images from a sharp image $S$ by ID-Blau.
That is, given a blur condition map $C=[x; y; z]$ and a sharp image $S$, ID-Blau can produce a blurred image $B = \textbf{ID-Blau}(S, C)$. Additionally, we create four blur condition maps based on $C$ for further demonstration, including unit horizontal or vertical blur orientations, $C^1=[1; 0; z]$ and $C^2=[0; 1; z]$, horizontally inverse $C$ as $C^3=[-x; y; z]$, and $C^4=[-x; y; 2z]$ with twice the magnitude for $C^3$. 

%
%
Besides, we have more examples to demonstrate the generalizability of ID-Blau by utilizing sharp images not in the training set to yield different blurred images in the supplementary material. 
These visuals attest that ID-Blau can effectively generate controllable and realistic blurred images. Thus, it can augment training data to improve image deblurring performance.

\section{Experiments}
\footnotetext{The authors from the universities in Taiwan completed the experiments on the datasets.}

\begin{table*}[t!]
\centering
\setlength{\tabcolsep}{0.1mm}
\caption{Evaluation results on the GoPro, HIDE, and RealBlur datasets, where ``Baseline'' and ``+ID-Blau'' denote the deblurring performances without and with ID-Blau, respectively.}
\vspace{-0.1in}
\begin{tabular}{cc|lc| lc| lc| lc| cc}
\hline\hline
& & \multicolumn{2}{c|}{\bf{GoPro}} & \multicolumn{2}{c|}{\bf{HIDE}} & \multicolumn{2}{c|}{\bf{RealBlur-J}} & \multicolumn{2}{c|}{\bf{RealBlur-R}} & \multicolumn{2}{c}{\bf{Average Gain}} \\
\multicolumn{2}{c|}{Model}  & PSNR   & SSIM   & PSNR  & SSIM & PSNR  & SSIM & PSNR  & SSIM & PSNR  & SSIM  \\ \hline
\multirow{2}{*}{MIMO-UNet+} & Baseline &   32.44    &  0.957   &  30.00   &   0.930 &   31.92    &  0.919   & 39.10  & 0.969  \\ & +ID-Blau &   \bf32.93 (+0.49)    &  \bf0.961   &  \bf30.68 (+0.68)  &   \bf0.938    &  \bf31.96 (+0.04)  &   \bf0.921 &  \bf39.38 (+0.28)  &   \bf0.971 & \bf+0.37 & \bf+0.004 \\ \hline

\multirow{2}{*}{Restormer} & Baseline &   32.92    &  0.961   &  31.22   &   0.942 &   32.88    &  0.933   &   40.15  & \bf0.974  \\ & +ID-Blau &   \bf33.51 (+0.59)   &  \bf0.965   &  \bf31.66 (+0.44)  &   \bf0.947 &   \bf33.11 (+0.23)   &  \bf0.937   &  \bf 40.31 (+0.16) &   \bf0.974 & \bf+0.36 & \bf+0.003 \\ \hline

\multirow{2}{*}{Stripformer} & Baseline &   33.08    &  0.962   &  31.03   &   0.940 &   32.48    &  0.929   &  39.84   &   0.974 \\ & +ID-Blau &   \bf33.66 (+0.58)   &  \bf0.966   &  \bf31.50 (+0.47)  &   \bf0.944 &   \bf33.77 (+1.29)   &  \bf0.940   &  \bf41.06 (+1.22)  &   \bf0.977 & \bf+0.89 & \bf+0.006 \\ \hline

\multirow{2}{*}{FFTformer} & Baseline &   34.21    &  0.969   &  31.62   &   0.946 &   32.62    &  0.933   &  40.11  &   0.973 \\ & +ID-Blau &   \bf34.36 (+0.15)   &  \bf0.970   &  \bf31.94 (+0.32)  &   \bf0.949 &  \bf32.88 (+0.26)   &  \bf0.934   &  \bf40.45 (+0.34)  &   \bf0.975 & \bf+0.27 & \bf+0.002 \\ \hline

\multicolumn{2}{c|}{Average Gain} & \bf+0.45  & \bf+0.003 & \bf+0.48 & \bf+0.005 & \bf+0.46 & \bf+0.005 & \bf+0.50 & \bf+0.002 & - & - \\ \hline

\hline\hline
\end{tabular}
\label{Tab:Performance}
\vspace{-0.2in}
\end{table*}

\subsection{Implementation Details}
\paragraph{ID-Blau.} 
We adopt the GoPro training set~\cite{Nah_2017_CVPR}, which contains $2,103$ blurred and sharp images for training ID-Blau. 
The training also requires the blur condition maps computed based on the blurred images.  
We use a batch size of $32$ and randomly crop images with the size of $128 \times 128$ for training $5,000$ epochs. 
We adopt the Adam optimizer and maintain a fixed learning rate of $1e^{-4}$.  
After optimizing ID-Blau, we utilize sharp images in the GoPro training set and create extra blur condition maps to produce additional $10,000$ $1280\times720$ blurred images by randomly modifying blur orientations and magnitudes based on the originally computed blur condition maps. 
%
%
The newly generated $10,000$ blurred and sharp pairs can largely enrich the GoPro training set to enhance the performance of a deblurring model. 
ID-Blau has 9.5 million parameters, and it takes 3.4 seconds to generate a $1280\times720$ blurred image on an Nvidia 3090 graphics card.

\vspace{-0.2in}
\paragraph{Deblurring models.} 
We adopt four prominent deblurring models, including MIMO-UNet+~\cite{MIMO}, Restormer~\cite{Zamir2021Restormer}, Stripformer~\cite{Tsai2022Stripformer}, and the state-of-the-art FFTformer~\cite{Kong_2023_CVPR} to demonstrate the effectiveness of ID-Blau. 
Following previous works, we evaluate deblurring performance on the GoPro, HIDE~\cite{su2017deep}, and RealBlur~\cite{rim_2020_ECCV} datasets.
The GoPro dataset comprises $2,103$ training pairs and $1,111$ testing pairs.
The HIDE dataset contains $2,025$ image pairs only for testing. 
The RealBlur dataset contains $3,758$ training pairs and $980$ testing pairs.  
We pre-train each deblurring model on the generated $10,000$ training pairs for $500$ epochs. After that, we fine-tune each deblurring model on the GoPro training set following its default training setting and then evaluate it on the GoPro testing set and the HIDE dataset.
Besides, we fine-tune each deblurring model on the RealBlur training set to demonstrate the effectiveness of ID-Blau on real-world blurred images. 
Since all compared methods utilized the GoPro and RealBlur training sets for evaluating the RealBlur testing set, ID-Blau ensures a fair comparison without relying on extra training data.

\subsection{Experimental Results}
\paragraph{Quantitative Results.} 
We compare the deblurring performance of four baselines and their ID-Blau-powered versions in Table~\ref{Tab:Performance}, where ``Baseline'' and ``+ID-Blau'' denote the deblurring performance without and with ID-Blau, respectively.      
Table~\ref{Tab:Performance} shows that ID-Blau significantly improves four prominent deblurring models, including MIMO-UNet+~\cite{MIMO}, Restormer~\cite{Zamir2021Restormer}, Stripformer~\cite{Tsai2022Stripformer}, and the state-of-the-art FFTformer~\cite{Kong_2023_CVPR}, on the GoPro, HIDE, and RealBlur datasets.
ID-Blau enhances those models' performance in PSNR by $0.45$dB, $0.48$dB, $0.46$dB, and $0.50$dB averagely on GoPro, HIDE, RealBlur-J, and RealBlur-R, respectively.
In particular, with ID-Blau, the state-of-the-art performance has been advanced to $34.36$dB ($+0.15$dB) on GoPro, $31.94$dB ($+0.32$dB) on HIDE, $33.77$dB ($+0.89$dB) on RealBlur-J, and $41.06$dB ($+0.72$dB) on RealBlur-R, where the enhanced  FFTformer achieves new SOTA results on GoPro and HIDE, and the enhanced Stripformer performs the best on ReaBlur-J and RealBlur-R. 
These comprehensive quantitative results demonstrate that ID-Blau significantly improves the performances of the four deblurring models on four widely-used datasets, showcasing ID-Blau's effectiveness and robustness as a data augmentation scheme for deblurring.

\begin{figure*}[t!]
\begin{center}
\includegraphics[width=1\textwidth]{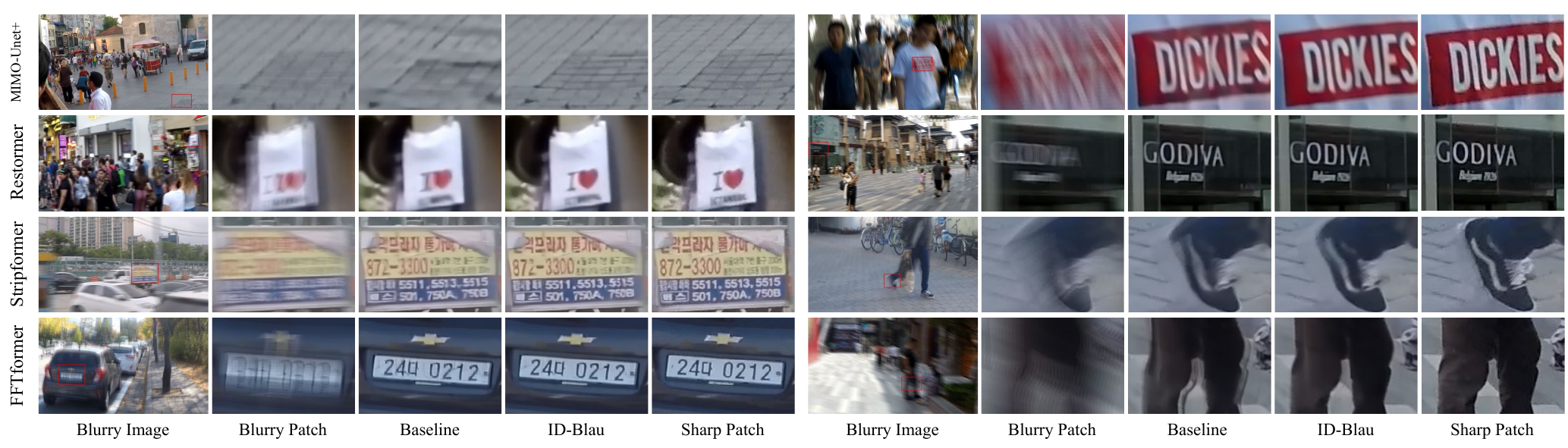}
\end{center}
\vspace{-0.2in}
\caption{Qualitative results on the GoPro testing set (left) and the HIDE dataset (right).
}
\label{fig:viz_Gopro_HIDE}
\vspace{-0.1in}
\end{figure*}

\begin{figure*}[t!]
\begin{center}
\includegraphics[width=1\textwidth]{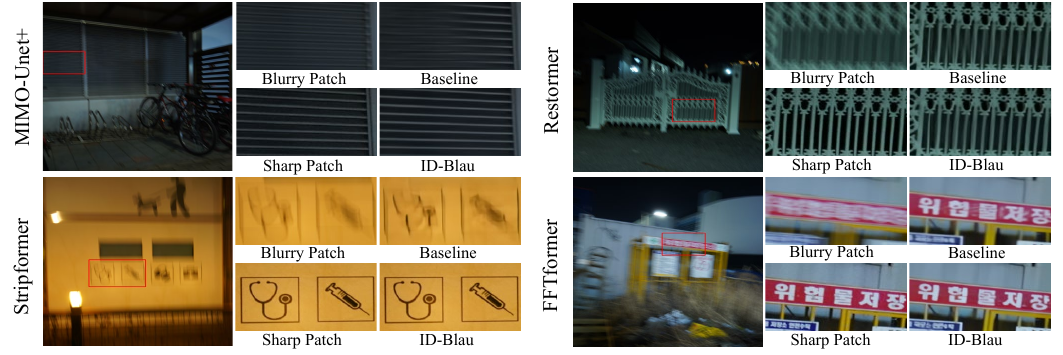}
\end{center}
\vspace{-0.2in}
\caption{Qualitative results on the RealBlur-J testing set.
}
\label{fig:viz_Realblur}
\vspace{-0.2in}
\end{figure*}

\vspace{-0.2in}
\paragraph{Qualitative Results.} 
Figure~\ref{fig:viz_Gopro_HIDE} compares the visual qualities of the results with the four deblurring baselines (denoted by ``Baseline'') and their ID-Blau-powered versions (denoted by ``+ID-Blau'') on GoPro and HIDE. 
The results show that ID-Blau significantly enhances deblurring visual qualities compared to the Baselines. Besides, we demonstrate visuals of deblurring models on RealBlur-J in Figure~\ref{fig:viz_Realblur}. 
We also demonstrate deblurring results on real-world blurry images~\cite{zhang2020deblurring} in the supplementary material.

\subsection{Ablation Studies}
We analyze the impact of ID-Blau on deblurring performance using GoPro. 
For efficiency, we train MIMO-UNet~\cite{MIMO}, a compact version of MIMO-UNet+, for $1,000$ epochs for analysis. 
If not specifically mentioned, ``Baseline'' denotes MIMO-UNet trained on the GoPro training set without using any additional augmented data samples.

\vspace{-0.2in}
\paragraph{Effect of Augmenting Blur Orientations.} 
Table~\ref{Tab:orientations} shows the effect of augmenting blur orientations with different angles on deblurring performance. 
To this end, with the original blur magnitudes, we rotate the blur orientations $[x; y]$ by four fixed angles ($30^\circ$, $60^\circ$, $90^\circ$, and $120^\circ$) plus a randomly selected angle for each produced blurred image (denoted ``Random"). 
The results indicate that blur augmentation with all specific/random angles achieves performance improvements similar to those of Baseline. 
This suggests no need to employ specific angles in blur condition maps for augmentation. 
For implementation convenience, we randomly alter blur orientations $[x; y]$ to $[-x; y]$, $[x; -y]$, or $[-x; -y]$ to augment blurred images with different orientations, achieving a similar performance improvement as well.

\begin{table}[t!]
\centering
\setlength{\tabcolsep}{1mm}
\caption{Impact of modifying blur orientations on deblurring performance in PSNR (dB).}
\vspace{-0.1in}
\begin{tabular}{l|cccccc}
\hline\hline
 & Baseline &  $30^\circ$  &  $60^\circ$  & $90^\circ$   & $120^\circ$ & Random \\ \hline
PSNR & 31.22 & 31.99   &  32.00  & 31.97 & 31.95  & 31.96  \\
\hline\hline
\end{tabular}
\vspace{-0.2in}
\label{Tab:orientations}
\end{table}

\begin{figure}
\begin{center}
\includegraphics[width=1\columnwidth]{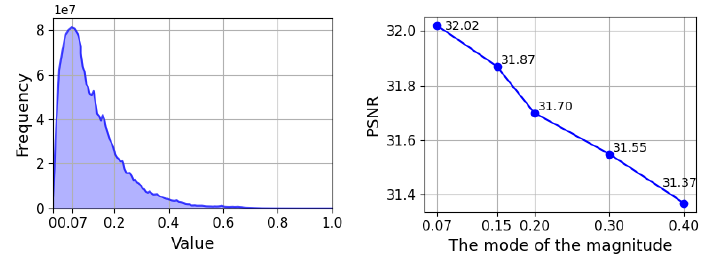}
\end{center}
\vspace{-0.3in}
\caption{Left: Distribution of blur magnitudes within the GoPro training set. Right: Illustration of the effect of altering the distribution's mode at different values on the deblurring performance.}
\label{fig:Ablation_fig1}
\vspace{-0.1in}
\end{figure}

\vspace{-0.2in}
\paragraph{Effect of Adjusting Blur Magnitudes.} 
Figure~\ref{fig:Ablation_fig1} illustrates the effect of modifying blur magnitudes on the deblurring performance. 
We first investigate the distribution of blur magnitudes within GoPro, in which most blurred images have blur magnitudes around with the peak at $0.07$.
Subsequently, we augment blurred images with different blur magnitudes by shifting the peak of magnitude distribution to different values, including $0.07$, $0.15$, $0.2$, $0.3$, and $0.4$. 
The results demonstrate that shifting the distribution's peak to a larger value causes decreased performance, indicating that generating blurred images that deviate from the original distribution of the dataset leads to a performance drop. 
Therefore, we modify blur magnitudes but keep the peak of blur magnitudes at around $0.07$.

\begin{figure}
\begin{center}
\includegraphics[width=1\columnwidth]{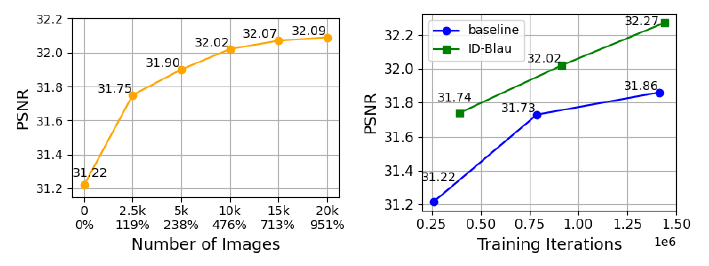}
\end{center}
\vspace{-0.25in}
\caption{Deblurring performance (PSNR) versus the number of augmented samples by ID-Blau (left) and versus the number of training iterations (right).}
\label{fig:ablation_training_iterations}
\vspace{-0.1in}
\end{figure}

\vspace{-0.2in}
\paragraph{Effect of the Number of Augmented Samples.} 
The left plot of Figure~\ref{fig:ablation_training_iterations} shows the impact of the number of augmented training pairs by ID-Blau on the deblurring performance. 
The performance gain of ID-Blau over that of the baseline, trained on GoPro without augmentation, increases from $0.53$dB to $0.87$dB with the number of augmented data samples (from $2,500$ to $20,000$), demonstrating ID-Blau's ability to improve the deblurring performance without the need of collecting additional training data samples. 
Augmenting $10,000$ additional samples (476\% of the original data size) achieves a reasonable tradeoff between the performance gain and training complexity. 
Since ID-Blau does not require extra training samples but consumes additional training complexity, we also investigate whether the performance improvements come from the additional training complexity. 
As shown in the right plot of Figure~\ref{fig:ablation_training_iterations}, ID-Blau stably outperforms the baseline with the same training complexities (\textit{i.e.}, the numbers of training iterations), where both the pre-training and fine-tuning process of ID-Blau are considered.  
Both the above experiments verify the performance gain with ID-Blau-based augmentation.  

\begin{figure}[t!]
\begin{center}
\includegraphics[width=1\columnwidth]{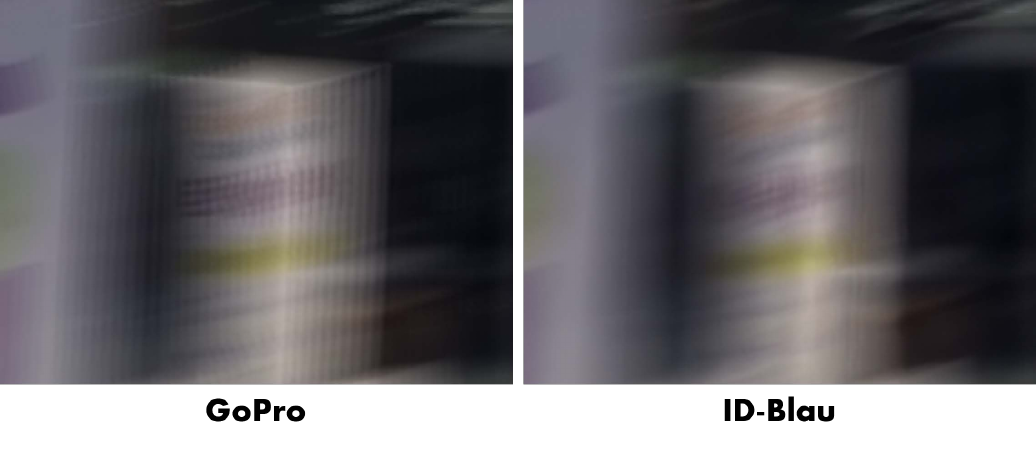}
\end{center}
\vspace{-0.2in}
\caption{Left: Blurry image in GoPro dataset. Right: Blurry image generated by ID-Blau. ID-Blau can generate blurry images without overlapped artifacts compared to the GoPro dataset.
}
\label{fig:compare_dataset}
\vspace{-0.2in}
\end{figure}

\vspace{-0.2in}
\paragraph{Effect of the Continuous Reblurring by ID-Blau.} 
Figure~\ref{fig:compare_dataset} shows two blurred images from GoPro and their reblurred versions by ID-Blau. 
Since GoPro's blurred images are generated by blur accumulation as in~(\ref{eqa:accumulaiton}), they contain unnaturally overlapping artifacts. 
In contrast, ID-Blau can generate similar blurred images by simulating continuous motion trajectories, yielding more realistic blurred images without such discontinuous overlapping artifacts. 
Therefore, ID-Blau can generate more realistic blurred images to improve the deblurring models' performances.

\vspace{-0.2in}
\paragraph{Performance Gain with a Diffusion Process.} In ID-Blau, we adopt the diffusion process in~\cite{NEURIPS2020_4c5bcfec} to generate blurred versions of sharp images with specified blur conditions. 
However, without the diffusion process, we can still utilize the same reblurring model to learn the mapping from a sharp image plus a computed blur condition map to the corresponding blurred image with the typical L1 loss function. 
%
%
To analyze the impact of ID-Blau with or without the diffusion process on deblurring performance, in Table~\ref{Tab:Ablation_model_base}, we compare Baseline (the same as in Figure~\ref{fig:ablation_training_iterations}), ``w/ ID-Blau$^\dag$," (ID-Blau without diffusion), and ``w/ ID-Blau'' (our model). 
%
The results show that ``w/ ID-Blau$^\dag$'' and ``w/ ID-Blau'' improve the Baseline by $0.57$dB and $0.80$dB, respectively, and the diffusion model achieves $0.23$dB additional performance gain.  

\begin{table}[t!]
\centering
\setlength{\tabcolsep}{1mm}
\caption{Effectiveness of using a diffusion model in ID-Blau on MIMO-UNet's performance in PSNR (dB), where  ``ID-Blau$^\dag$'' denotes the proposed ID-Blau without its diffusion process.}
\vspace{-0.1in}
\begin{tabular}{l|ccc}
\hline\hline
Model & Baseline & w/ ID-Blau$^\dag$   & w/ ID-Blau  \\ \hline
PSNR & 31.22 & 31.79 & 32.02  \\
\hline\hline
\end{tabular}
\vspace{-0.1in}
\label{Tab:Ablation_model_base}
\end{table}

\begin{table}[t!]
\centering
\setlength{\tabcolsep}{1mm}
\caption{Comparison between ID-Blau and BGAN regarding the DBGAN's performance in PSNR (dB),  where ID-Blau$^\dag$ denotes the proposed ID-Blau without its diffusion process.}
\vspace{-0.1in}
\begin{tabular}{l|cccc}
\hline\hline
Model  & Baseline   & w/ BGAN & w/ ID-Blau$^\dag$ & w/ ID-Blau     \\ \hline
PSNR & 30.43 & 30.92  & 31.57 & \bf31.67  \\
\hline\hline
\end{tabular}
\label{Tab:DBGAN_compare}
\vspace{-0.2in}
\end{table}

\vspace{-0.2in}
\paragraph{Comparison between ID-Blau- and BGAN-based Augmentation.}
Besides, we compare the performance gains of ID-Blau and that of BGAN~\cite{zhang2020deblurring}, a representative reblurring GAN for augmenting blurred images to improve the DBGAN deblurring model. 
To fairly compare ID-Blau with BGAN, we utilize ID-Blau to improve DBGAN's deblurring performance. 
In Table~\ref{Tab:DBGAN_compare}, ``Baseline'' denotes DBGAN's performance trained on GoPro, and ``w/ BGAN'' denotes DBGAN's performance with BGAN-augmented data. ``w/ ID-Blau$^\dag$'' and ``w/ ID-Blau'' denote DBGAN's performance using ID-Blau-augmented data without and with the diffusion process, respectively.
Note that GAN-based methods tend to generate unpredictable and uncontrollable blurred images, whereas ID-Blau can generate controllable results based on specified blur conditions. 
As a result, ``w/ ID-Blau$^\dag$'' and ``w/ ID-Blau'' improve the PSNR performance of DBGAN by $1.14$dB and $1.24$dB. 
Thanks to the proposed continuous blur condition map with ID-Blau, the simplified ``w/ ID-Blau$^\dag$'' without the diffusion process can still offer an additional gain of $0.47$dB over that with ``w/ BGAN.'' 
This verifies the effectiveness of ID-Blau compared to the GAN-based reblurring model BGAN.
\section{Conclusion}
We proposed a diffusion-based reblurring model that can take a sharp image and a controllable pixel-wise blur condition map to synthesize a blurred image. 
To train the reblurring model, we parameterized the blur patterns of a blurred image with their orientations and magnitudes and implicitly represented them in a continuous blur condition field. 
With the model, we presented ID-Blau, an effective data augmentation scheme for image deblurring, where we sample various blur conditions from the field to produce diverse, realistic blurred images and enrich the training set. 
Experimental results have shown that ID-Blau can significantly improve the performance of state-of-the-art deblurring models.
\section{Acknowledgments}
This work was supported in part by the National Science and Technology Council (NSTC) under grants 112-2221-E-A49-090-MY3, 111-2628-E-A49-025-MY3, 112-2634-F-002-005, 112-2221-E-004-005, and 113-2923-E-A49-003-MY2. 
This work was funded in part by Qualcomm through a Taiwan University Research Collaboration Project.

{
    \small
    \bibliographystyle{ieeenat_fullname}
    \bibliography{main}
}


\end{document}